%% file: arxiv.tex
\documentclass[letterpaper]{article} %
\usepackage{aaai24}  %
\usepackage{times}  %
\usepackage{helvet}  %
\usepackage{courier}  %
\usepackage[hyphens]{url}  %
\usepackage{graphicx} %
\usepackage{natbib}  %
\usepackage{caption} %
\usepackage{algorithm}
\usepackage{algorithmic}

\usepackage{amssymb}
\usepackage{amsmath}
\usepackage{multirow}
\usepackage{caption}
\usepackage{booktabs}
\usepackage{subcaption}
\usepackage{appendix}

\usepackage{newfloat}
\usepackage{listings}
\usepackage{pifont} %
\usepackage[table,dvipsnames]{xcolor}   
\DeclareCaptionStyle{ruled}{labelfont=normalfont,labelsep=colon,strut=off} %
\floatstyle{ruled}
\newfloat{listing}{tb}{lst}{}
\floatname{listing}{Listing}
\newcommand{\cmark}{\textcolor{blue}{\ding{51}}}
\newcommand{\xmark}{\textcolor{red}{\ding{55}}}

\nocopyright 

\title{IncreLoRA: Incremental Parameter Allocation Method for \\Parameter-Efficient Fine-tuning}

\newcommand{\meetyou}{\raisebox{3pt}{\includegraphics[scale=0.050]{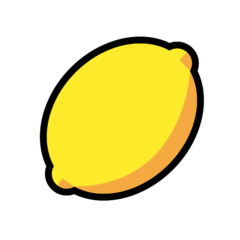}}}
\newcommand{\osaka}{\raisebox{3pt}{\includegraphics[scale=0.055]{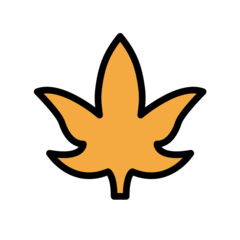}}}
\newcommand{\astar}{\raisebox{3pt}{\includegraphics[scale=0.050]{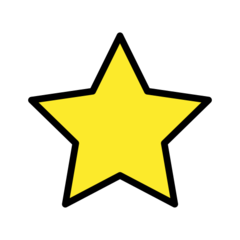}}}
\newcommand{\atom}{\raisebox{3pt}{\includegraphics[scale=0.02]{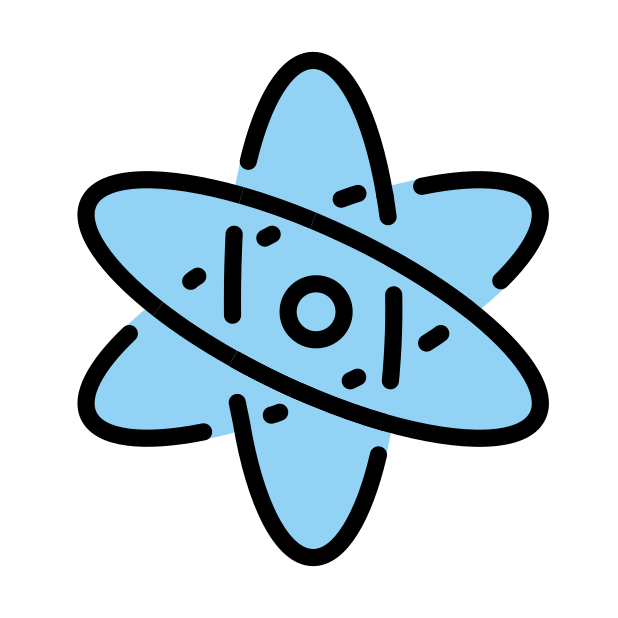}}}
\newcommand{\email}{\raisebox{3pt}{\includegraphics[scale=0.02]{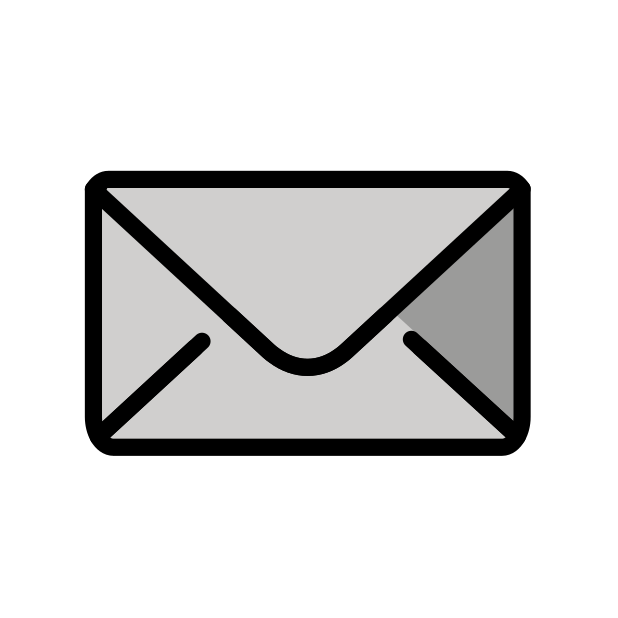}}}
\newcommand{\emailmain}{\raisebox{-3.5pt}{\includegraphics[scale=0.02]{icon/email.png}}}

\newcommand{\meetyoumain}{\raisebox{-2pt}{\includegraphics[scale=0.050]{icon/lemon.png}}}
\newcommand{\osakamain}{\raisebox{-2pt}{\includegraphics[scale=0.055]{icon/leaf.png}}}
\newcommand{\astarmain}{\raisebox{-2pt}{\includegraphics[scale=0.050]{icon/star.png}}}
\newcommand{\atommain}{\raisebox{-2pt}{\includegraphics[scale=0.02]{icon/atom.png}}}

\newcommand\blfootnote[1]{%
  \begingroup
  \renewcommand\thefootnote{}\footnote{#1}%
  \addtocounter{footnote}{-1}%
  \endgroup
}

\author{
\normalfont{\textbf{Feiyu Zhang}\meetyou\atom, \textbf{Liangzhi Li}\meetyou\email, \textbf{Junhao Chen}\meetyou, \\
\textbf{Zhouqiang Jiang}\meetyou, \textbf{Bowen Wang}\osaka\meetyou, \textbf{Yiming Qian}\astar\meetyou }\\
\meetyoumain{Meetyou AI Lab}, \osakamain{Osaka University}\\
\atommain{University of Electronic Science and Technology of China (UESTC)}\\
\astarmain{Agency for Science, Technology and Research (A*STAR)}\\
\{\texttt{zhangfeiyu1}, \texttt{liliangzhi}, \texttt{chenjunhao}, \texttt{jiangzhouqiang}\}\texttt{@xiaoyouzi.com},\\
\texttt{bowen.wang@is.ids.osaka-u.ac.jp}, \texttt{qiany@ihpc.a-star.edu.sg}
}

\usepackage{bibentry}

\begin{document}

\maketitle
\blfootnote{\emailmain Corresponding author.}

\begin{abstract}
With the increasing size of pre-trained language models (PLMs), fine-tuning all the parameters in the model is not efficient, especially when there are a large number of downstream tasks, which incur significant training and storage costs. Many parameter-efficient fine-tuning (PEFT) approaches have been proposed, 
among which, Low-Rank Adaptation (LoRA) is a representative approach that injects trainable rank decomposition matrices into every target module. Yet LoRA ignores the importance of parameters in different modules. To address this problem, many works have been proposed to prune the  parameters of LoRA. However, under limited training conditions, the upper bound of the rank of the pruned parameter matrix is still affected by the preset values. We, therefore, propose IncreLoRA, an incremental parameter allocation method that adaptively adds trainable parameters during training based on the importance scores of each module. This approach is different from the pruning method as it is not limited by the initial number of training parameters, and each parameter matrix has a higher rank upper bound for the same training overhead. We conduct extensive experiments on GLUE to demonstrate the effectiveness of IncreLoRA. The results show that our method owns higher parameter efficiency, especially when under the low-resource settings where our method significantly outperforms the baselines. Our code is publicly available.\footnote{https://github.com/FeiyuZhang98/IncreLoRA}
\end{abstract}

\section{Introduction}

\begin{figure}[t!]
    \centering
    \includegraphics[width=0.4\textwidth]{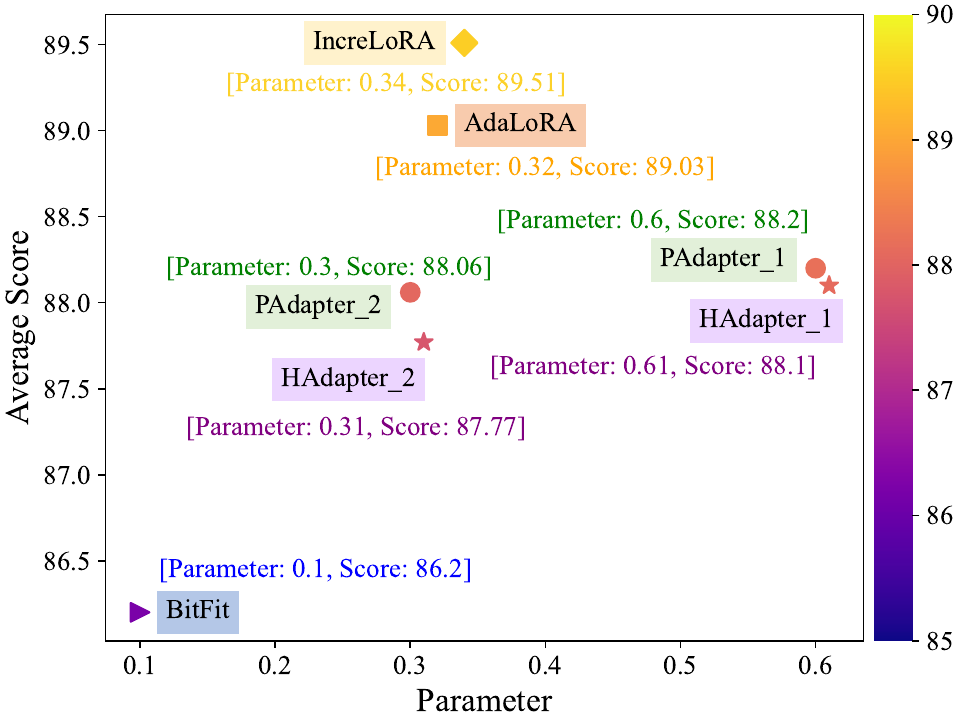}
    \caption{Fine-tuning results for different methods and parameter budget on the GLUE benchmark, all experiments are based on DeBERTaV3-base. We compare our method with BitFit \cite{peft:bitfit}, PAdapter \cite{peft:PAdapter}, HAdapter \cite{peft:adapter}, and AdaLoRA \cite{peft:lora}, with the x-axis representing the number of parameters (M), and the y-axis representing the average score (Avg). Our approach achieves a better trade-off between efficiency and performance.}
    \label{fig:score_dot}
\end{figure}

In recent years, pre-training models with large amounts of data and fine-tuning them for different downstream tasks have become one of the most successful training paradigms \cite{plm:1,plm:2,plm:3,plm:4,plm:5,plm:6}. However, with the increasing size of popular pre-trained models, such as BERT (110M$\sim$340M) \cite{plm:2}, T5 (60M$\sim$11B) \cite{plm:6} , GPT3 (175B) \cite{plm:7}, LLaMA (7B$\sim$65B) \cite{plm:8}, it becomes difficult to train these models due to graphics card memory limitations, and it is expensive to train and store the full parameters of the model for each downstream task. 

Parameter-Efficient Fine-Tuning (PEFT) is a promising solution to the above problems. As the name implies, parameter-efficient fine-tuning achieves performance comparable to full parametric fine-tuning by making only a small number of parameter changes to the model. Due to its broad application prospects, researchers have proposed various methods for efficient parameter tuning \cite{peft:adapter,peft:prompt,peft:prefix,peft:bitfit,peft:lora}. Some methods only fine-tune part of the parameters of the original model \cite{peft:diffpurning,peft:bitfit}, while others train a batch of new parameters and add them to the original model \cite{peft:adapter,peft:PAdapter,peft:lst}. LoRA \cite{peft:lora} is one of the representative methods, which simulates a low-rank update matrix $\Delta W$ of the same size as the original weight matrix through the product of two low-rank matrices, and updates only two low-rank matrices during the fine-tuning process (usually only $0.01\%-0.5\%$ of the original matrix parameter count, depending on the rank of the matrices), but achieves a full update of the original weight matrix. Despite the low-rank parameter update $\Delta W$, experimental results show that the fine-tuning performance of LoRA approaches or exceeds that of full fine-tuning \cite{peft:lora}.

\begin{figure}[tp!]
    \centering
    \includegraphics[width=0.4\textwidth]{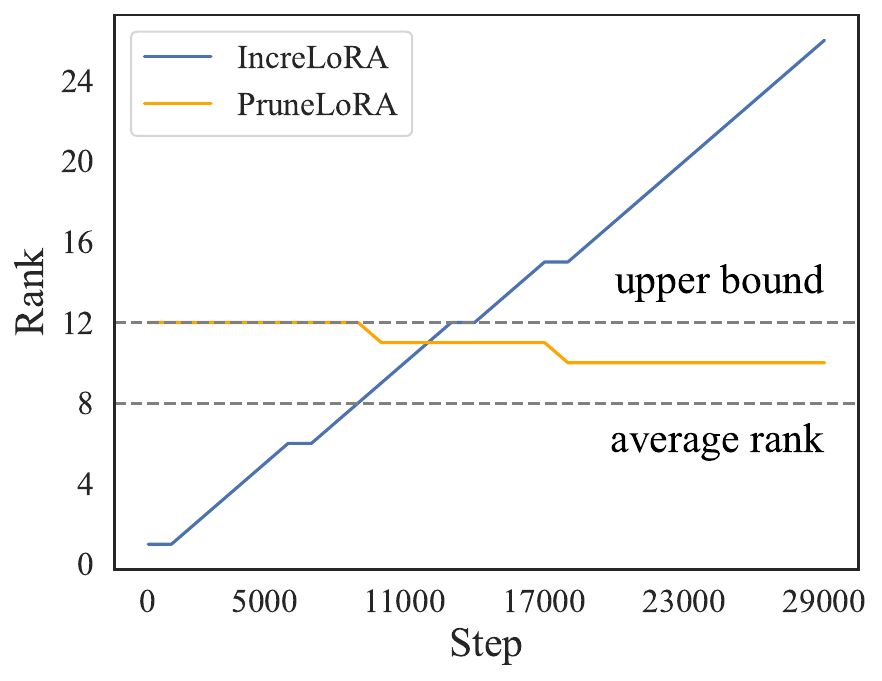}
    \caption{We illustrate the variations in rank of \texttt{layer.10.attention.self.value\_proj} when fine-tuning DeBERTaV3-base on MNLI, using IncreLoRA and Pruning LoRA methods respectively.}
    \label{fig:rank_plot}
\end{figure}

LoRA has been introduced to various tasks \cite{c:1,c:2,c:3} due to its simple structure, but it sets the same rank for each update matrix equally, ignoring the differences between modules, and it has been demonstrated that assigning different ranks to different update matrices achieves better fine-tuning performance for the same total number of trainable parameters \cite{peft:adalora, c:2}. This type of method can be recognized as a pruning of the LoRA method, although better performance has been achieved, however, the rank of each LoRA module is still limited by the rank before pruning (The module rank here actually refers to the rank of the update matrix obtained from the product of two low-rank matrices in the module, in the following, we will use both terms). The aforementioned structured pruning methods typically set the initial average rank to be 1.5 times that of the final state. For instance, if the average rank after pruning is 8, then the initial rank for each module would be 12. As illustrated in Figure \ref{fig:rank_plot}, the rank of unimportant modules will decrease (potentially to 0, meaning no parameter updates) after the pruning process, and more important modules have a rank that is most likely greater than 8, but still cannot exceed the initial upper bound of rank=12. In order to make each module have a higher upper bound on the rank after pruning, it is necessary to set each module with a higher initial rank, which will undoubtedly raise the training cost of the model and require more iterations to make the training stable. 

To resolve the above contradiction, we propose IncreLoRA, an incremental trainable parameter allocation method, which automatically increases the rank based on the importance score of each module during the training process, and each module has a higher rank upper bound at the same training cost. In order to prevent insufficient training of subsequently added parameters, we use a new pre-training technique to allow the parameters added to the module to have a favorable initial state, which we call advance learning to show the difference. Moreover, we assign independent learning rate curves to these parameters to ensure the stability of training. As shown in Figure \ref{fig:score_dot}, our experimental results show that our proposed method achieves better performance, especially in the low-resource case, and our method performs well enough to match other methods with higher parameter budgets. The contributions of this paper are summarized as follows:

\begin{itemize}
\item We propose IncreLoRA which has a lower training cost and higher rank upper bound in comparison to pruning methods. To the best of our knowledge, it is the first parameter-efficient fine-tuning method that incrementally assigns trainable parameters.
\item We introduce a new parameter pre-training method, which allows the parameters to find a better state in advance while ensuring training stability and without taking up additional training time.
\item We conduct extensive experiments on natural language understanding benchmark, and the results demonstrate the effectiveness of the model we introduce.
\end{itemize}

\section{Related Work}

As a method to reduce the cost of fine-tuning and storage of large-scale pre-trained models, PEFT has received more and more attention from researchers in recent years, with different methods varying in terms of memory efficiency, storage efficiency, and inference overhead. According to whether the original parameters are fine-tuned in the training phase, existing PEFT methods can be categorized into selective and additive.

\subsection{Selective Method}
Selective Methods pick and update the model based on its original parameters. Fine-tuning only a few top layers of the network \cite{c:4} can be seen as an early attempt at PEFT, while many methods proposed in recent years select specific model layers \cite{c:9} or internal modules, e.g., BitFit \cite{peft:bitfit} updates only the bias parameters in the model, which dramatically reduces the number of trainable parameters, but this brute-force approach can only lead to sub-optimal performance. Unstructured parameter fine-tuning based on the scoring function in parameter selection of trainable parameters \cite{c:5,c:6,c:7,c:8} is a more complete solution. FishMask \cite{c:10} performs multiple iterations on the sub-data according to the model and calculates the amount of parameter Fisher information, and then constructs a mask to select the $k$ parameters with the maximum Fisher information for updating. However, since existing deep learning frameworks and hardware cannot well support the computation of such unstructured sparse matrices, it has a similar memory footprint during training and full parameter fine-tuning.
\subsection{Additive Method}
The Additive Method replaces full-parameter fine-tuning by adding trainable additional parameters to the prototype network, Adapters are trainable mini-modules inserted into the original network, which were first applied to multi-domain image categorization \cite{c:11}, and later introduced after Transformer's Attention and FNN layers \cite{c:12}, and have given rise to many variants \cite{c:13,c:14,c:15}. Unlike methods like Adapters, LoRA \cite{peft:lora} simulates the update of the weight matrix in the model by the product of two low-rank matrices, and the trained parameters can be added to those in the original network during the inference phase without incurring additional inference overhead. Prefix-Tuning \cite{peft:prefix} adds the trainable parameters before the sequence of hidden states in all layers. A similar approach is Prompt-Tuning \cite{c:16}. LST \cite{peft:lst} feeds hidden states from the original network into a small trainable ladder side network via shortcut connections, so that the gradients do not need to be backpropagated through the backbone network, further reducing the memory utilization.
\subsection{Hybrid Method}
Based on the impressive performance of PEFT methods mentioned above, many works have focused on mining the effectiveness of different methods and designing unified frameworks to combine them for optimal performance \cite{c:17,c:18,c:19}, however, they still belong to the early manual attempts. Recently, many works have been based on the idea that the parameter redundancy has also existed in PEFT modules, pruning the trainable parameters to achieve better fine-tuning performance with a limited parameter budget \cite{c:20,c:21,c:2}. \citet{c:2} applies both structured and unstructured methods for pruning LoRA. \citet{c:21} lets all layers share the same single PEFT module (e.g., Adapter, LoRA, and Prefix-Tuning) and learns a binary mask for each layer to select different sub-networks.

\section{Methodology}
The idea of our approach is the gradual increase of the trainable parameters of the model during the training process. Based on our observations, incremental parameter allocation without any improvement strategies is likely to result in training insufficiency and instability, thereby leading to poor fine-tuning performance. Note that we focus on how to improve this particular training approach instead of proposing a new importance scoring function. For this Section, we first present how we reconstruct the low-rank matrices in LoRA in Section 3.1, then present the details of the parameter allocation strategy in Section 3.2, and eventually describe how to utilize the reconstructed low-rank matrices for advance learning in Section 3.3.

\subsection{Reconstructing Low-Rank Matrices}
For a pre-trained weight matrix $W^{\left( 0 \right)}\in \mathbb{R}^{out\times in}$, we define the parameter efficient fine-tuning as an update matrix $\Delta W$. LoRA \cite{peft:lora} decomposes it into the product of two low-rank matrices:

\begin{figure}[tp!]
    \centering
    \includegraphics[width=0.48\textwidth]{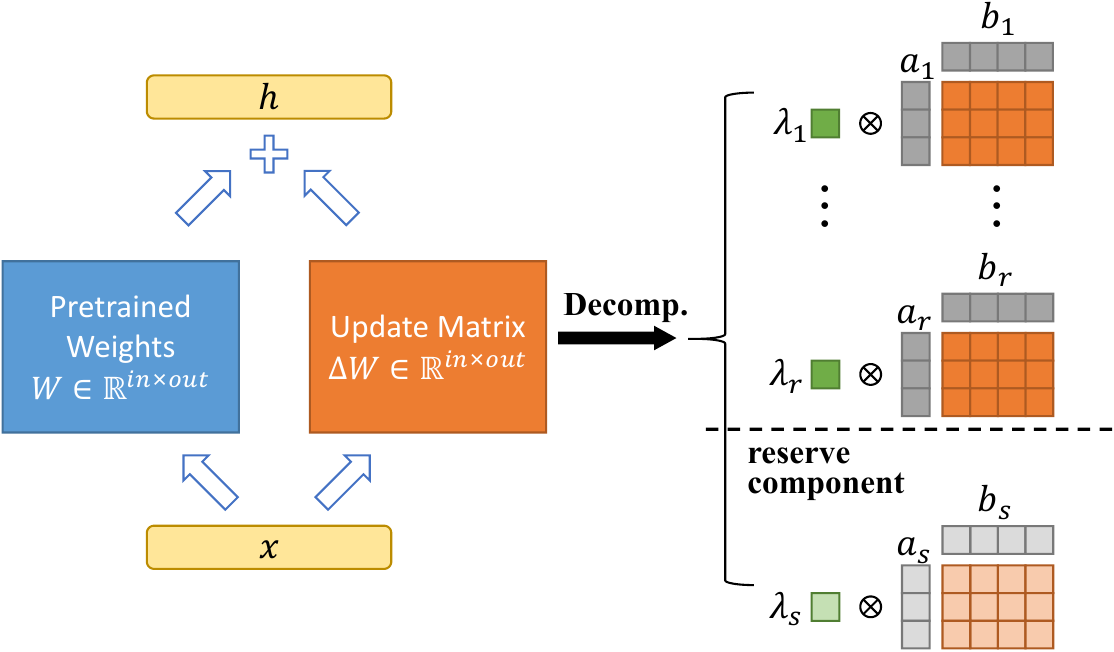}
    \caption{An illustration of the low-rank adapters in our model, where $x$ is the input of each module and $h$ is the output of the module. The update matrix $\Delta W$ can be decomposed into $r$ components and an additional preparatory component. where $\lambda_s$ is fixed to 1e-5 and the remaining parameters are trainable.}
    \label{fig:lora_figure}
\end{figure}

\begin{equation}
\label{eq:1}
W=W^{\left( 0 \right)}+ \Delta W=W^{\left( 0 \right)}+BA ,
\end{equation}
where $ A \in\mathbb{R}^{r\times in}, B \in\mathbb{R}^{out\times r} $ and the rank $r\ll \min \left( in, out \right)$. A scaling factor is also applied to this update matrix, usually set to $1/r$. $A$ and $B$ can be viewed as a combination of a set of vectors: $A=\left[ a_1,a_2,\cdots ,a_r \right], B=\left[ b_1,b_2,\cdots ,b_r \right]$, where $a_i \in \mathbb{R}^{in}, b_i \in \mathbb{R}^{out}$. Thus equation (1) can be further disassembled as:

\begin{equation}
\begin{split}
W &= W^{\left( 0 \right)}+w_1+w_2+\cdots+w_r \\
  &= W^{\left( 0 \right)}+b_1a_1+b_2a_2+\cdots+b_ra_r ,
\end{split}
\end{equation}
where $w_i$ is a matrix of rank 1 obtained by the product of the vectors $b_i, a_i$ . 

To match our proposed advance learning strategy, we add a scaling $\lambda_i$ to each $w_i$, so equation~(\ref{eq:1}) is updated into:

\begin{equation}
W = W^{\left( 0 \right)}+\sum_{i=1}^r{\lambda_i w_i} = W^{\left( 0 \right)}+\sum_{i=1}^r{\lambda_i b_i a_i} ,
\end{equation}
where $\lambda_i$ is updated through backpropagation. We apply random Gaussian initialization to each $a_i$ and $b_i$, while $\lambda_i$ is initialized to zero to ensure that the initial state of $\Delta W$ is zero. In order to reduce the dependency between $w_i$ and to ensure the stability of training, we apply a regularization term which is introduced by~\citet{peft:adalora}:

\begin{equation} 
R\left( A,B \right) =\lVert A^TA-I \rVert _{\text{F}}^{2}+\lVert B^TB-I \rVert _{\text{F}}^{2} ,
\end{equation}  
where $I$ is an identity matrix, this forces $A$ and $B$ to be orthogonal after training, and the Frobenius norm of the trained $w_i$ equal to one, and the effect of each $w_i$ on $\Delta W$ controlled by $\lambda_i$. Let $\Lambda = \left[\lambda_1,\lambda_2,\cdots,\lambda_r \right]$, then $\Delta W = B\Lambda A$, which is similar in form to the singular value decomposition (SVD), except that $\lambda_i$ can be an arbitrary constant, whereas the singular values in the SVD are non-negative. In our model, we replace the original LoRA matrix with a SVD-like triplets, and since $\Lambda$ is a diagonal matrix, we can store it with only a one-dimensional tensor, which introduces only $r$ additional parameters for each module.

In the training of common transformer-base models, LoRA is typically appended to the key and value projection matrices in the attention layer. In our approach, however, parameter updates are applied to all linear layers, with the update matrix ranks being scheduled by a allocator according to importance scores.

\subsection{Incremental Parameter Allocation}
Unlike the pruning approach, the amount of trainable parameters for our model increases with training. We utilize rank to denote the parameter budget for each module, and the sum of all ranks at the end of training, denoted as $r^{final}$, represents the total number of ranks we can assign. At model initialization, the rank of each module is one. During the parameter allocation phase, the total rank of the model, $r^{total}$, increases linearly until the end of parameter allocation, when $r^{total}$ equals $r^{final}$.

At the t$-th$ step of the phase, an importance score $S^{\left( t \right)}_k$ needs to be maintained for each module, for $k=1,2,...,n$, where $k$ is the index of the module and $n$ is the total number of modules. We make $S^{\left( t \right)}_k$ equal to the average of the scores of all parameters of the update matrix \cite{c:22}:

\begin{equation}
\label{eq:5}
S^{\left( t \right)}_k = avg\left( \left| \Delta W_k * grad\left( \Delta W_k \right) \right| \right) .
\end{equation}

The update matrix is modeled by the product of $B$, $\Lambda$ and $A$, we can can capture the gradient when backpropagating to this matrix via a backward hook. Since we take mini-batch training, the score of each step is affected by the randomness of sampling, so we improve the reliability of the score by sensitivity smoothing and uncertainty quantification \cite{c:23}:
\begin{align}
\label{eq:6}
I_{k}^{\left( t \right)}   & =\beta_1 I_{k}^{\left( t-1 \right)}+\left( 1-\beta _1 \right) S_{k}^{\left( t \right)}, \\
\label{eq:7}
U_{k}^{\left( t \right)}   & =\beta_2 U_{k}^{\left( t-1 \right)}+\left( 1-\beta_2 \right) \left| I_{k}^{\left( t \right)} - S_{k}^{\left( t \right)}\right|, \\
\label{eq:8}
\hat{S}_{k}^{\left( t \right)} & =I_{k}^{\left( t \right)}*U_{k}^{\left( t \right)},
\end{align}
where $0<\beta _1,\beta _2<1$. $\hat{S}_{k}^{\left( t \right)}$ is the importance score we actually utilize. At intervals of $\nu$ steps, the allocator adds a new set of parameters for all modules with scores in the top-$h$, where $\nu$ and $h$ are hyperparameters, we default $\nu$ to equal warmup steps. Theoretically, at the start of each training session, the upper bound of rank for each module is $r^{final} / h$.

\begin{algorithm}[ht!]
\caption{IncreLoRA. $\mathcal{D}$ is dataset, $\mathcal{M}$ is target module set, $\mathcal{A},\mathcal{B},\mathcal{E}$ is the set of parameter matrices, $T$ is total steps, $\mathcal{W}$ is warmup steps, $\eta$ is learning rate, $h$ is the number of modules selected in each round, $r^{final}$ is final total rank.}
\label{alg:algorithm}
\textbf{Input}:$\mathcal{D}$, $T$, $\mathcal{W}$, $\eta$, $h$, $r^{final}$\\
\textbf{Parameter}: $\mathcal{A},\mathcal{B},\mathcal{E}$
\begin{algorithmic}[1] %
\FOR{$M_k$ in $\mathcal{M}$} 
    \STATE Append $a_k,b_k,\lambda _k$ to $A_k^{\left( 0 \right)}, B_k^{\left( 0 \right)}, \Lambda _k^{\left( 0 \right)}$ for advance learning
    \STATE init\_normal$\left( a_k,b_k \right)$; freezed$\left( \lambda _k \right)\gets1e-5 $ 
\ENDFOR
\FOR{$t = 1,...,T$} 
    \IF {$ r^{total} < r^{final}$}
        \STATE Compute $\hat{S}_{k}^{\left( t \right)}$ by (\ref{eq:5}),(\ref{eq:6}),(\ref{eq:7}),(\ref{eq:8}) \textbf{for} $M_k$ \textbf{in} $\mathcal{M}$
        \IF {$ t\ \%\ \mathcal{W} \ ==\ 0 $}
            \FOR{$\hat{S}_{k}^{\left( t \right)}$ in top$-h$ \{$\hat{S}_{1}^{\left( t \right)},...,\hat{S}_{n}^{\left( t \right)}$\} } 
                \STATE Activate old $\lambda_k$
                \STATE Append new $a_k,b_k,\lambda_k$ to $A_k^{\left( t \right)}, B_k^{\left( t \right)}, \Lambda _k^{\left( t \right)}$
                \STATE init\_normal$\left( a_k,b_k \right)$; freezed$\left( \lambda _k \right)\gets1e-5 $ 
                \STATE Assign new learning rate scheduler for $a_k$, $b_k$, and old $\lambda_k$
            \ENDFOR
        \ENDIF
    \
    \ELSE
    \STATE Mask out reserve components \textbf{for} $M_k$ \textbf{in} $\mathcal{M}$
    \ENDIF
    \STATE Update $\mathcal{A}^{\left( t \right)},\mathcal{B}^{\left( t \right)},\mathcal{E}^{\left( t \right)}$
\ENDFOR
\end{algorithmic}
\textbf{Output}: The fine-tuned parameters \{$\mathcal{A}^{\left( T \right)},\mathcal{B}^{\left( T \right)},\mathcal{E}^{\left( T \right)}$\} \\
\end{algorithm}

\begin{table*}[ht!]
\large
\centering
\resizebox{\textwidth}{!}{
\begin{tabular}{l|c|ccccccccc}
\toprule
\multirow{2}{*}{\textbf{Method}} & \multirow{2}{*}{\textbf{\#Params}} & \textbf{MNLI}  & \textbf{SST-2} & \textbf{CoLA}  & \textbf{QQP}   & \textbf{QNLI}  & \textbf{RTE}   & \textbf{MRPC}     & \textbf{STS-B} & \textbf{ALL} \\
                                 &                                    & Acc.            & Acc.            & Mcc.            & Acc.            & Acc.            & Acc.            & Acc.               & Corr.           & Avg.         \\ \midrule
Full FT                          & 184M                               & 89.90          & 95.63          & 69.19          & 92.40 & 94.03          & 83.75          & 89.46             & 91.60          & 88.25        \\ \midrule
BitFit                           & 0.1M                               & 89.37          & 94.84          & 66.96          & 88.41          & 92.24          & 78.70          & 87.75             & 91.35          & 86.20        \\ \midrule
HAdapter                         & 1.22M                              & 90.13          & 95.53          & 68.64          & 91.91          & 94.11          & 84.48          & 89.95             & 91.48          & 88.28        \\
PAdapter                         & 1.18M                              & 90.33          & 95.61          & 68.77          & 92.04          & 94.29          & 85.20          & 89.46             & 91.54          & 88.41        \\
LoRA$_{r=8}$                     & 1.33M                              & 90.65          & 94.95          & 69.82          & 91.99          & 93.87          & 85.20          & 89.95             & 91.60          & 88.50        \\
AdaLoRA                          & 1.27M                              & 90.76          & 96.10          & 71.45          & 92.23          & \textbf{94.55}          & 88.09          & 90.69             & 91.84          & 89.46        \\
IncreLoRA                        & 1.33M                              & \textbf{90.93$\pm$0.04} & \textbf{96.21$\pm$0.20} & \textbf{71.82$\pm$0.76} & \textbf{92.25$\pm$0.03} & 94.45$\pm$0.12   & \textbf{88.21$\pm$0.21} & \textbf{91.01$\pm$0.62}    & \textbf{91.93$\pm$0.26} & \textbf{89.61$\pm$0.16}    \\ \midrule
HAdapter                         & 0.61M                              & 90.12          & 95.30          & 67.87          & 91.65          & 93.76          & 85.56          & 89.22             & 91.30          & 88.10        \\
PAdapter                         & 0.60M                              & 90.15          & 95.53          & 69.48          & 91.62          & 93.98          & 84.12          & 89.22             & 91.52          & 88.20        \\
HAdapter                         & 0.31M                              & 90.10          & 95.41          & 67.65          & 91.54          & 93.52          & 83.39          & 89.25             & 91.31          & 87.77        \\
PAdapter                         & 0.30M                              & 89.89          & 94.72          & 69.06          & 91.40          & 93.87          & 84.48          & 89.71             & 91.38          & 88.06        \\
LoRA$_{r=2}$                     & 0.33M                              & 90.30          & 94.95          & 68.71          & 91.61          & 94.03          & 85.56          & 89.71             & 91.68          & 88.32        \\
AdaLoRA                          & 0.32M                              & 90.66          & 95.80          & 70.04          & 91.78          & 94.49          & 87.36          & 90.44             & 91.63          & 89.03        \\
IncreLoRA                        & 0.34M                              & \textbf{90.71$\pm$0.05} & \textbf{96.26$\pm$0.13} & \textbf{71.13$\pm$0.77} & \textbf{91.78$\pm$0.09} & \textbf{94.65$\pm$0.16} & \textbf{88.52$\pm$0.30}    & \textbf{91.13$\pm$0.98} & \textbf{91.89$\pm$0.21} & \textbf{89.51$\pm$0.22}  \\ \bottomrule
\end{tabular}
}
\caption{Experiment results based on the GLUE development set. We report Matthews correlation for CoLA and average correlation for STS-B, as well as the accuracy of the other datasets. We report baseline results directly from \citet{peft:adalora}. Best performances are highlighted in bold. We run the experiment on 5 different random seeds and report the mean and standard deviation.}
\label{tab:main_result}
\end{table*}

\subsection{Advance Learning}
As described in section 3.1, the update matrix $\Delta W_k$ can be viewed as a weighted summation of all the components $w_{k,i}$ with Frobenius norm 1. All the values in $w_{k,i}$ can be viewed as a state, and the weight $\lambda_{k,i}$ is viewed as the influence of this state on $\Delta W_k$. We want the added $w_{k,i}$ to have a good state every time we increase the rank of the update matrix. Therefore, when we initialize the model and increase the module's rank, we add an extra reserve component $w_{k,s}$ (i.e., $a_{k,s}, b_{k,s}$) and $\lambda_s$, where $a_{k,s}$ and $b_{k,s}$ are initialized with random Gaussians and can be trained, and $\lambda_{k,s}$ is initialized to 1e-5 and fixed. $w_{k,s}$ will find a better state with training, and since $\lambda_{k,s}$ is a very small value, $w_{k,s}$ will not have any practical inference on the model. In our experiments, we found that the order of magnitude of $\lambda_{k,i}$ generally lies between 1e-1 and 1e-2, so that $\lambda_{k,s}$ is a relatively small value and $w_{k,s}$ does not significantly affect the model. When the module needs to increase the rank, the $\lambda_{k,s}$ will be activated and $w_{k,s}$ will gradually take effect with subsequent training.

Furthermore, applying the same learning rate curve to all trainable parameters is not suitable, because each time a new set of parameters is added for advance learning, these parameters are in a randomized state and cannot follow the learning rate of other parameters that have been trained with a favorable state. Therefore, we set new learning rate curves for these parameters, which restart warmup and drop the learning rate to zero at the end of training like all other parameters. We summarize the detailed process of IncreLoRA in Algorithm 1.

\section{Experiment}
In this section, we first evaluate the performance of our proposed IncreLoRA through extensive experiments on general natural language processing benchmarks. Subsequently, we focus on demonstrating the effectiveness of the individual components through ablation experiments, as well as analyzing the reasons for the superior performance of incrementally assigning trainable parameters in low-resource scenarios.

\begin{table*}[htb!]
\centering
\begin{tabular}{@{}ccccccc@{}}
\toprule
\textbf{orthogonal regularization}   & \textbf{advance learning}   & \textbf{restart warmup}   & \textbf{\#Params} & \textbf{RTE} & \textbf{\#Params} & \textbf{STS-B} \\ \midrule
\xmark                    & \xmark           & \xmark         & 0.27M             & 85.92        & 0.32M             & 91.64          \\
\xmark                    & \cmark       & \xmark         & 0.36M             & 87.00        & 0.25M             & 91.89          \\
\cmark                & \xmark           & \xmark         & 0.34M             & 87.73        & 0.34M             & 91.78          \\
\cmark                & \cmark       & \xmark         & 0.31M             & 88.09        & 0.34M             & 91.88          \\
\cmark                & \xmark       & \cmark         & 0.34M             & 87.86        & 0.34M             & 91.85          \\
\cmark                & \cmark       & \cmark     & 0.33M             & 88.81        & 0.36M             & 92.07          \\ \bottomrule
\end{tabular}
\caption{Ablation experiments on IncreLoRA, we report experimental results and number of parameters on RTE and STS-B. Each of these components leads to performance improvements and their combination is key to the success of our approach.}
\label{tab:ablation}
\end{table*}

\subsection{Experimental Setting}
\subsubsection{Datasets}
GLUE \cite{e:1} is a generalized natural language understanding assessment benchmark that includes a variety of tasks such as sentence relationship recognition, sentiment analysis, and natural language reasoning, from which we select eight tasks for systematic evaluation, including Corpus of Linguistic Acceptability (CoLA), Multi-Genre Natural Language Inference (MNLI), Microsoft Research Paraphrase Corpus (MRPC), Question Natural Language Inference (QNLI), Quora Question Pairs (QQP), Recognizing Textual Entailment (RTE), Stanford Sentiment Treebank (SST-2), Semantic Textual Similarity Benchmark (STS-B). The language of all datasets is English (BCP-47 en).

\subsubsection{Baselines}
We compare our methods to Full Fine-tuning, BitFit \cite{peft:bitfit}, HAapter \cite{peft:adapter}, PAdapter \cite{peft:PAdapter}, LoRA \cite{peft:lora} and AdaLoRA \cite{peft:adalora}. BitFit only fine-tunes the bias in the model. HAdapter inserts the adapter after the attention module and FNN module in the Transformer, and PAdapter only inserts the adapter after the FNN module and LayerNorm. AdaLoRA is a structured pruning method applied to LoRA.

\subsubsection{Training Cost}
Disregarding the common pre-trained model weights and focusing solely on the differences among various methods. Assuming that all methods use the Adamw optimizer, each increase in the rank of the LoRA module requires $m$ parameters, then LoRA needs to occupy $3mr$ of memory. For structured pruning LoRA, assuming that the pruning ratio is 50\%, it needs to occupy $6mr$ of memory. Since our method needs to add a preparatory component for each module, it needs to occupy $3m(r+1)$ memory.

\subsubsection{Implementation Details}
We use the Pytorch \cite{e:5} and Transformers \cite{e:6} libraries to facilitate the construction of our experimental code, and all experiments are performed on NVIDIA 4090 GPUs. We implement IncreLoRA for fine-tuning DeBERTaV3-base \cite{e:3}, which has 12 layers with a hidden size of 768 and contains a total of 183 million parameters. We apply the update matrix to all weight matrices in the backbone network and adjust the total number of trainable parameters by the final total rank $r^{final}$ of all update matrices in the model, e.g., the number of trainable parameters in DeBERTav3-base is about 0.32M when the target average rank (which we denote as $r^{avg}$) is 2. We choose $r^{avg}$ from \{2, 4, 8, 12, 16, 24\}, which corresponds to $r^{total}$ of \{144, 288, 576, 864, 1152, 1728\} in DeBERTav3-base. We perform experiments with the settings $r^{avg} = 2$ and $r^{avg} = 8$ on all datasets of GLUE, and on several of them for all $r^{avg}$ levels. However, in practice, the real parameter total is not fixed. The total amount of parameters will be affected by different rank distributions because of the different sizes of the individual modules. In addition, the optimal checkpoint for some tasks occurs before the end of the parameter distribution, when only a lower-than-budgeted parameter count is required. Specific values and hyperparameter settings will be given in the appendix. 

\subsection{Experiment Result}
We compare the performance of IncreLoRA with the baseline model under different parameter budgets in Table \ref{tab:main_result}, and since the allocation of trainable parameters for our method is not fixed across tasks, we show the average number of parameters for all tasks. It can be seen that our method achieves comparable or better performance than the baseline model in different tasks with different parameter budgets. Moreover, on some datasets, IncreLoRA outperforms other methods that have four times the parameter budget. For example, when the parameter budget is 0.3M, our method achieves 96.26\% accuracy on SST-2, which is higher than full fine-tuning and AdaLoRA (1.27M). And on RTE, our method achieves 88.52\% accuracy, which is 3.61\% higher than full fine-tuning and 0.43\% higher than AdaLoRA (1.27M). 

In addition, we also find that IncreLoRA at the same budget level actually occupies more parameters, due to the fact that our approach prefers to add parameters to the FNN modules for most of the tasks, and in DeBERTaV3-base these modules add four multiples of the number of parameters at a time than the other modules. Nonetheless, the performance of our method is still remarkable, with an average score on GLUE that exceeds that of other higher-budget methods.

\begin{figure}[htp!]
    \centering
    \includegraphics[width=0.48\textwidth]{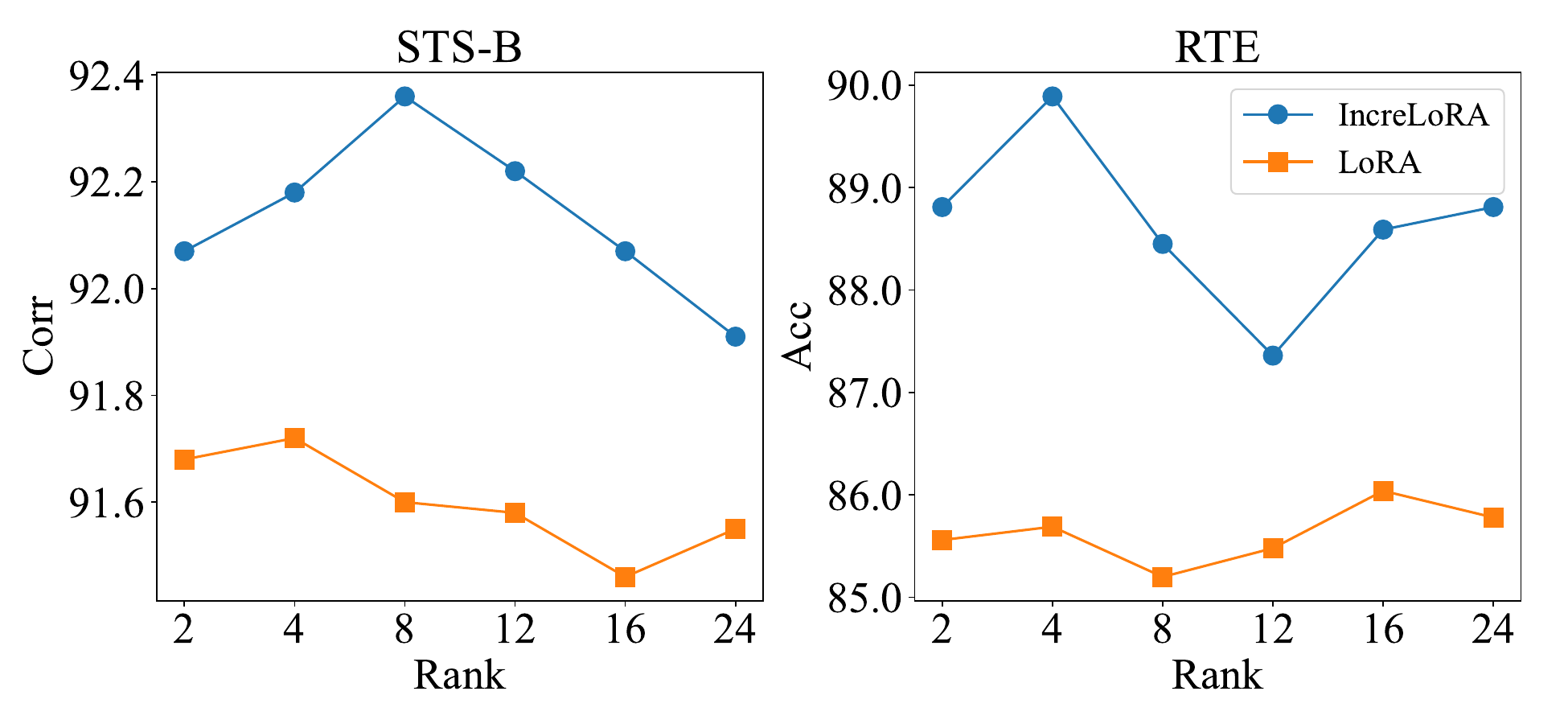}
    \caption{Fine-tuning performance under different parameter budgets. The x-axis represents the average rank, and the y-axis is the evaluation index of different data sets. Set the same learning rate under different parameter budgets.}
    \label{fig:diff_level}
\end{figure}

\subsection{Different Budget Levels}
We test the performance of IncreLoRA fine-tuning DeBERTaV3-base under different parameter budget levels and present the results in Figure \ref{fig:diff_level}. We find that on both the RTE and STS-B datasets, our method achieves significant performance improvements at different budget levels. And IncreLoRA requires only the lowest parameter budget and already exceeds the performance of LoRA for all parameter budget levels. For example, IncreLoRA achieves 88.81\% accuracy on RTE when rank equals 2, which is 2.77\% higher than the best performance of LoRA (86.04\%). This shows that our method has good parameter efficiency in low resource situations.

\subsection{Ablation Experiment}
In order to verify the effectiveness of the individual components, we perform ablation experiments in the setting of average rank = 2 and display the results in Table \ref{tab:ablation} and Figure \ref{fig:rte_stsb}. In this section, we focus on the interactions between the components. Note that orthogonal regularization is not our proposed method, but we still include it in the ablation experiments in order to verify its interactions with other components.
\subsubsection{Raw Incremental Parameter Allocation}
We eliminate all components and simply add trainable parameters dynamically during training, this process is the opposite of structured pruning based on LoRA and produces worse performance. Since the parameters added halfway through the process are not adequately trained, the performance of the model drops dramatically, even lower than the original LoRA.

\subsubsection{Advance Learning and Orthogonal Regularization}
Adding advance learning and orthogonal regularization individually to the original incremental parameter allocation both significantly improves the training performance of the model. On RTE, orthogonal regularization and advance learning bring 1.81\% and 1.08\% accuracy improvement, respectively.  The focus here on the effect of our proposed advance learning on orthogonal regularization. It can be seen that by adding advance learning to the orthogonal regular term, the accuracy of RTE and the average correlation of STS-B increase by 0.36\% and 0.10\%, respectively. \citet{peft:adalora} argues that keeping $A$ and $B$ orthogonal improves the parameter efficiency of the model. 

\begin{figure}[htp!]
    \centering
    \includegraphics[width=0.48\textwidth]{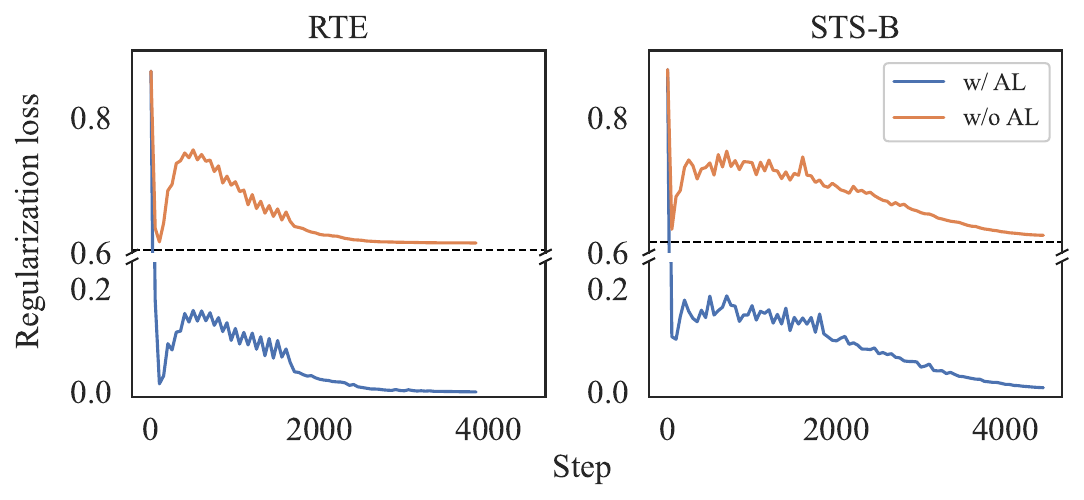}
    \caption{Regularization loss with and without advance learning. To save space, the middle part that does not contain critical information is truncated.}
    \label{fig:rte_stsb}
\end{figure}

As shown in Figure \ref{fig:rte_stsb}, $A,B$ still has a large orthogonal regularization loss after training, while after adding advance learning, the orthogonal regularization loss decreases rapidly, and $A,B$ has practically orthogonality. We believe this is due to the trade-off between predictive and regularization losses. When advance learning is removed, the added parameter gradient comes more from the prediction loss, which tends to prevent the regularization loss from decreasing. During advance learning, $\lambda _s$ is fixed to a smaller value, which limits the influence of the preparatory component on the model, the gradient comes more from the regularization loss, and when the preparatory component is activated, it tends to already have better orthogonality properties, which ensures the parametric efficiency of our method.

\subsubsection{Restart Warmup}
Although learning in advance allows $w_i$ to be activated with a better initial state, when the randomly initialized $a_i$ and $b_i$ follow the original learning rate scheduler, it is prone to unstable training due to the large variance of the mini-batch data distribution\cite{e:7}. When restarting warmup for each added parameter group, the performance of the model is significantly improved.

\subsection{Analysis}
\begin{figure}[hbp!]
\centering

\begin{subfigure}{0.5\textwidth}
\centering
\includegraphics[width=\textwidth]{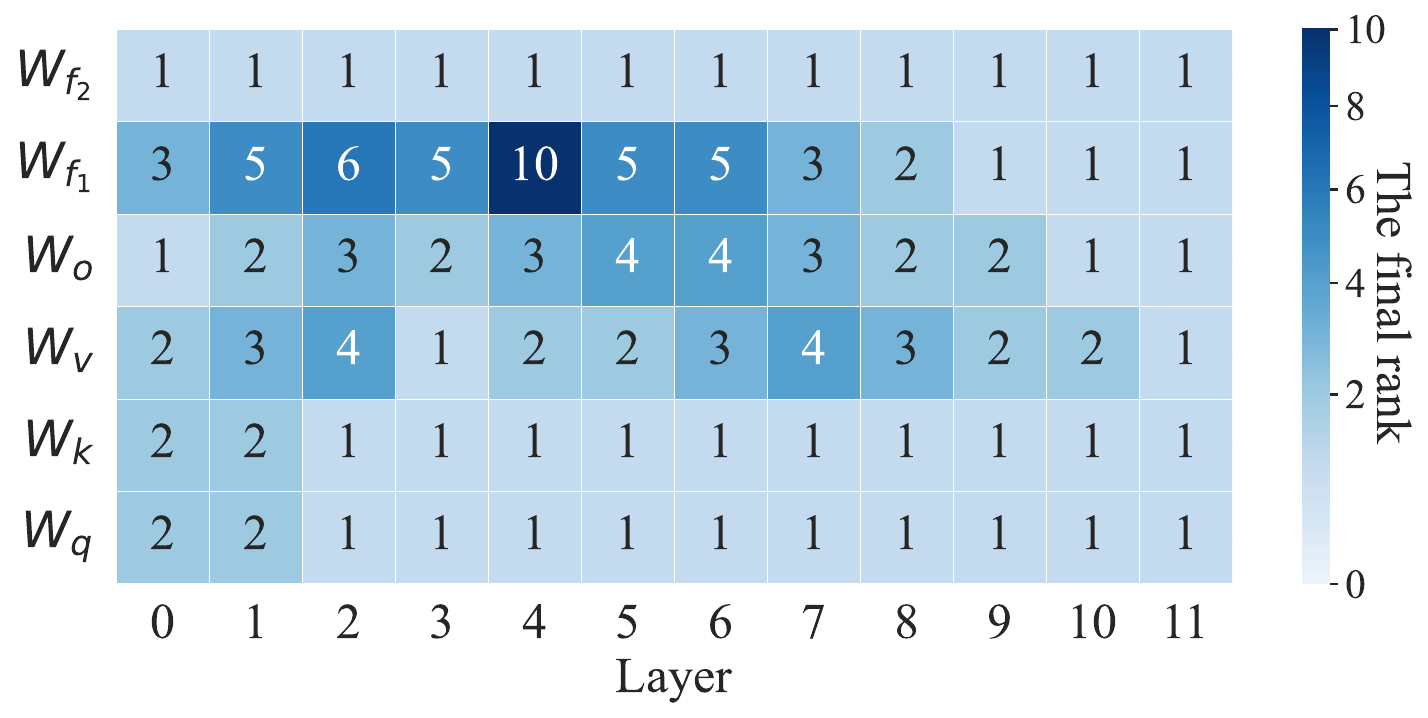}
\caption{IncreLoRA}
\end{subfigure}

\vspace{3mm} %

\begin{subfigure}{0.5\textwidth}
\centering
\includegraphics[width=\textwidth]{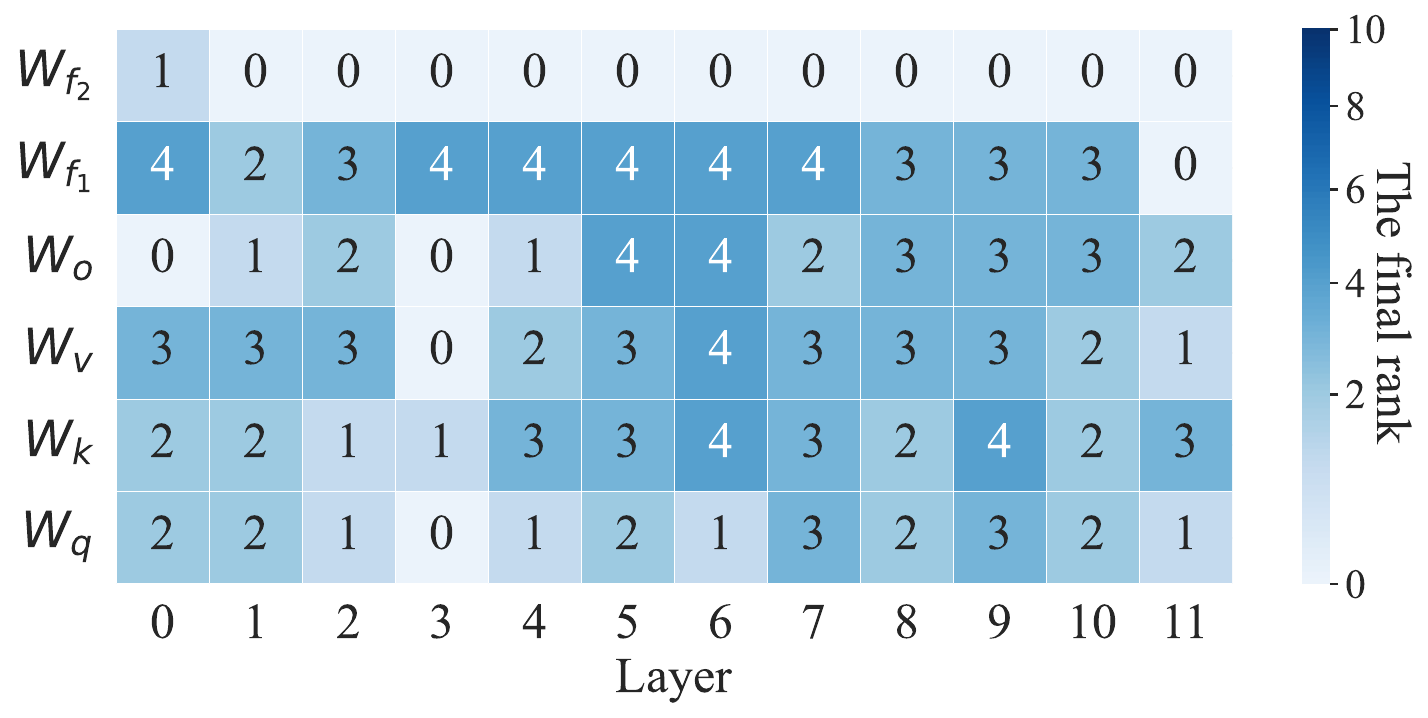}
\caption{AdaLoRA}
\end{subfigure}

\caption{The rank of each update matrix after fine-tuning DeBERTaV3-base on SST-2 using IncreLoRA and AdaLoRA, respectively. Where the horizontal coordinate represents the layer index and the vertical coordinate represents the type of module, from bottom to top, query/key/value projection, output projection in the self-attention, and two weight matrices in two-layer FFNs. }
\label{fig:rank analysis}
\end{figure}

In order to verify the validity of the high rank upper bound due to incremental parameter allocations, we investigate the impact of the rank distributions of IncreLoRA and AdaLoRA on the model performance. Specifically, we apply the two methods based on the same base model and parameter scales on the SST-2 respectively and save the rank distributions of their respective individual modules. As shown in Figure \ref{fig:rank analysis}, both methods favor assigning more rank to $W_v,W_o$ and $W_{f1}$, but IncreLoRA produces a more centralized distribution of rank. We hypothesize that this is one of the factors that make our approach fruitful on low resources,  as it ensures that important modules still have a high rank in low resource situations.

\begin{table}[h!]
\small
\centering
\begin{tabular}{l|cc|cc}
\toprule
\multirow{2}{*}{\textbf{Method}} & \multirow{2}{*}{\textbf{\#Params}} & \textbf{SST-2} & \multirow{2}{*}{\textbf{\#Params}} & \textbf{CoLA} \\
                                 &                                    & Acc.            &                                    & Acc.           \\ \midrule
IncreLoRA                        & 0.35M                              & 96.22          & 0.32M                              & 71.48         \\
AdaLoRA                          & 0.31M                              & 95.86          & 0.31M                              & 70.64         \\ \midrule
LoRA (r = Incre)                & 0.35M                              & 95.87          & 0.32M                              & 70.66         \\
LoRA (r = Ada)                  & 0.31M                              & 95.76          & 0.31M                              & 70.31         \\ \bottomrule
\end{tabular}
\caption{The rank distributions generated by IncreLoRA and AdaLoRA are used to initialize and retrain the two LoRAs separately. Average rank is 2.}
\label{tab:rank analysis}
\end{table}

We initialize the two new LoRA models with the two different rank distributions, and in order to minimize the experimental variability, we retain the SVD-like parameter matrix form and regularization terms common to both methods and train them with the same learning rate and number of epochs. As shown in Table \ref{tab:rank analysis}, the rank distribution obtained by IncreLoRA results in a model with better parameter efficiency. In addition, we find that the performance of the LoRA models with two different rank distributions is not as well as that of IncreLoRA and AdaLoRA, which suggests that the generated rank distributions and our proposed incremental parameter allocation method have some dependence, and only by dynamic adjusting the individual modules during the training process can the optimal performance be achieved.

\section{Conclusion}
In this paper, we propose a simple and effective parameter-efficient trimming method, IncreLoRA, for further improving the parameter efficiency of low-rank adapters. Unlike the structured pruning method, we incrementally assign trainable parameters during the training process. In addition, we reconstruct the low-rank matrix in LoRA and propose to utilize early learning and restarting warmup to improve the training effect and stability. Experimental results demonstrate the effectiveness of our proposed model and show superior performance with low resources.

\bibliography{aaai24}

\include{Appendix}

\end{document}

%% file: Appendix.tex
\onecolumn

\begin{appendices}
\section{GLUE DATASET STATISTICS}

We present the summary of the GLUE \cite{e:1} benchmark in the table below, where WNLI is not included in the experiment.

\begin{table*}[htp!]
\centering
\resizebox{0.8\textwidth}{!}{
\begin{tabular}{@{}lrrlll@{}}
\toprule
\textbf{Corpus} & \multicolumn{1}{l}{\textbf{Train}} & \multicolumn{1}{l}{\textbf{Dev}} & \textbf{Task}       & \textbf{Metrics}             & \textbf{Domain}     \\ \midrule
\multicolumn{6}{c}{Single-Sentence Task}                                                                                                                           \\ \midrule
CoLA            & 8.5k                               & 1k                               & acceptability       & Matthews corr.               & misc.               \\
SST-2           & 67k                                & 872                              & sentiment           & Acc.                         & movie reviews       \\ \midrule
\multicolumn{6}{c}{Similarity and Paraphrase Tasks}                                                                                                                \\ \midrule
MRPC            & 3.7k                               & 408                              & paraphrase          & Acc./F1                      & news                \\
STS-B           & 7k                                 & 1.5k                             & sentence similarity & Pearson/Spearman corr.       & misc.               \\
QQP             & 364k                               & 40k                              & paraphrase          & Acc./F1                      & social QA questions \\ \midrule
\multicolumn{6}{c}{Inference Tasks}                                                                                                                                \\ \midrule
MNLI            & 393k                               & 20k                              & NLI                 & Matched Acc./Mismatched Acc. & misc.               \\
QNLI            & 105k                               & 5.7k                             & QA/NLI              & Acc.                         & Wikipedia           \\
RTE             & 2.5K                               & 276                              & NLI                 & Acc.                         & news,Wikipedia      \\ 
WNLI            & 634                                & 71                               & coreference/NLI     & Acc.                         & fiction books      \\ \bottomrule
\end{tabular}
}
\caption{Task descriptions and statistics of GLUE. All tasks are single sentence or sentence pair classification, except STS-B, which is a regression task. MNLI has three classes; all other classification tasks have two.}
\end{table*}

\section{TRAINING DETAILS}
In all tasks, batch size, $\beta_1$, $\beta_2$ in equation (6), (7) are \{32, 0.85, 0.85\}. we set the allocation interval $\nu$ by default to the same value as warmup steps.
\begin{table*}[h]
\centering
\begin{tabular}{@{}ccccccc@{}}
\toprule
Dataset        & learning rate       & batch size & epochs & top\_h  & warmup steps & dropout rate \\ \midrule
\textbf{MNLI}  & $3.5\times 10^{-4}$ & 32         & 9      & 2       & 1000         & 0.2     \\
\textbf{SST-2} & $8\times 10^{-4}$   & 32         & 24     & 5       & 1000         & 0.1     \\
\textbf{CoLA}  & $8\times 10^{-4}$   & 32         & 25     & 3       & 100          & 0       \\
\textbf{QQP}   & $4\times 10^{-4}$   & 32         & 7      & 3       & 1000         & 0       \\
\textbf{QNLI}  & $7\times 10^{-4}$   & 32         & 5      & 12      & 500          & 0       \\
\textbf{RTE}   & $1.2\times 10^{-3}$ & 32         & 50     & 5       & 100          & 0       \\
\textbf{MRPC}  & $1\times 10^{-3}$   & 32         & 30     & 6       & 100          & 0.3     \\
\textbf{STS-B} & $2.2\times 10^{-3}$ & 32         & 25     & 4       & 100          & 0       \\ \bottomrule
\end{tabular}
\end{table*}

\section{QUESTION ANSWERING TASK}
We evaluate the performance of IncreLoRA on a Q\&A dataset with the pre-trained model DeBERTaV3-base. SQuADv2.0 is available at https://rajpurkar.github.io/SQuAD-explorer/.
\begin{table*}[ht]
\centering
\resizebox{0.4\textwidth}{!}{
\begin{tabular}{@{}l|ccc@{}}
\toprule
          & \multicolumn{3}{c}{\textbf{SQuADv2.0}}                             \\ \midrule
Full FT   & \multicolumn{3}{c}{85.4/88.4}                                      \\ \midrule
\# Params & 0.16\%               & 0.32\%               & 0.65\%               \\ \midrule
HAdapter  & 84.3/87.3            & 84.9/87.9            & 85.4/88.3            \\
PAdapter  & 84.5/87.6            & 84.9/87.8            & 84.5/87.5            \\
LoRA      & 83.6/86.7            & 84.5/87.4            & 85.0/88.0            \\
AdaLoRA   & 85.7/88.8            & 85.5/88.6            & 86.0/88.9            \\ \midrule
IncreLoRA & \textbf{86.05/88.81} & \textbf{85.89/88.84} & \textbf{86.17/89.07} \\ \bottomrule
\end{tabular}
}
\caption{Results with DeBERTaV3-base on SQuADv2.0. Here Params is the number of trainable parameters relative to that in full fine-tuning. We report EM/F1. The best results in each setting are shown in bold.}
\end{table*}

\clearpage

\section{EXPERIMENTAL STATISTICS}
We fine-tune DeBERTaV3-base on eight datasets using IncreLoRA and present the statistical results of the scaling ($\lambda$) for all module components ($w$) in the figure below, where all weights are taken as absolute values. It can be observed that most singular values fall within the range of 1e-1 to 1e-2.
\begin{figure*}[htp!]
    \centering
    \includegraphics[width=0.8\textwidth]{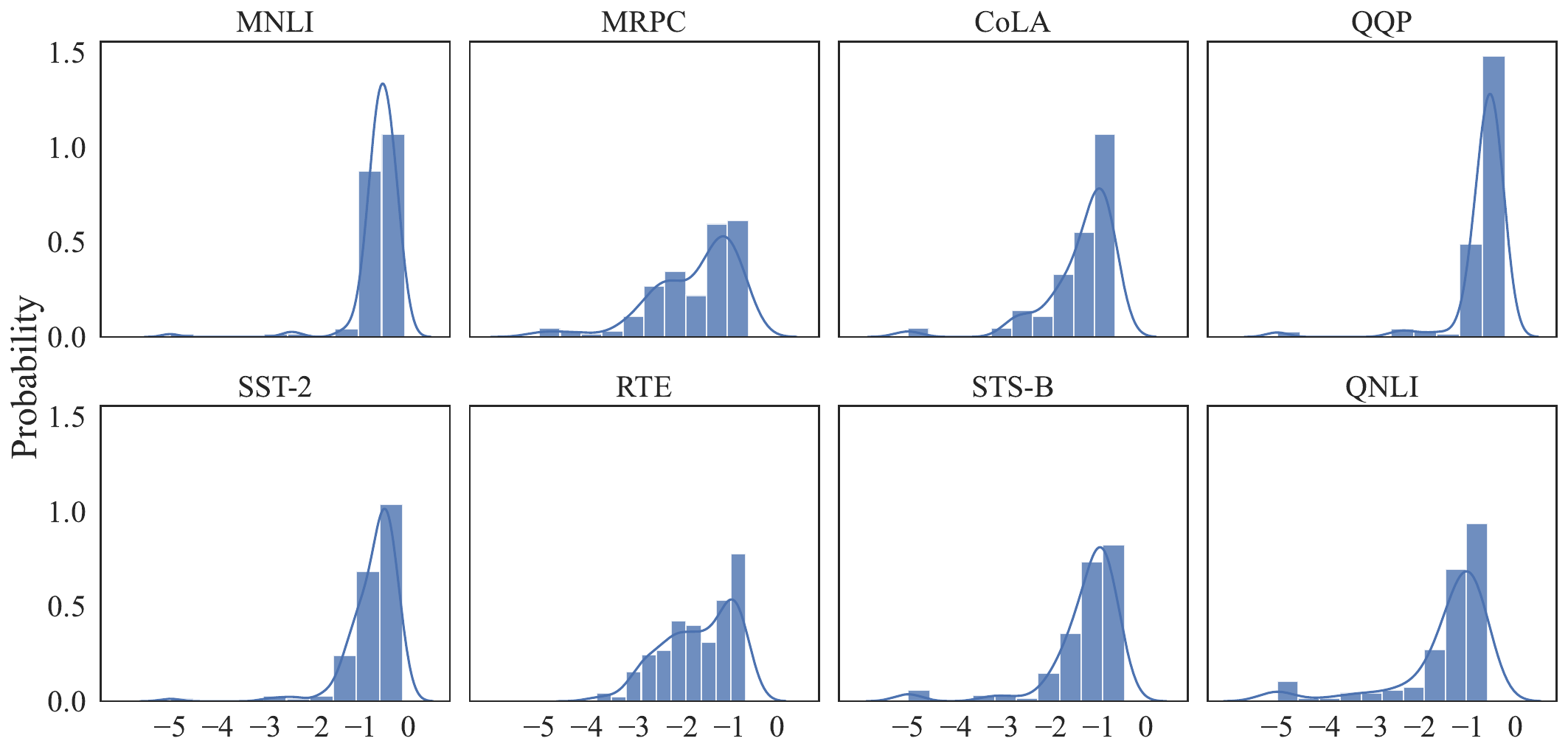}
    \caption{Scaling factor statistic plots. The horizontal axis represents the power of the scaling factor, the vertical axis represents the probability density.}
    \label{fig:lambda_statistic}
\end{figure*}

\begin{figure*}[hp!]
    \centering
    \includegraphics[width=0.8\textwidth]{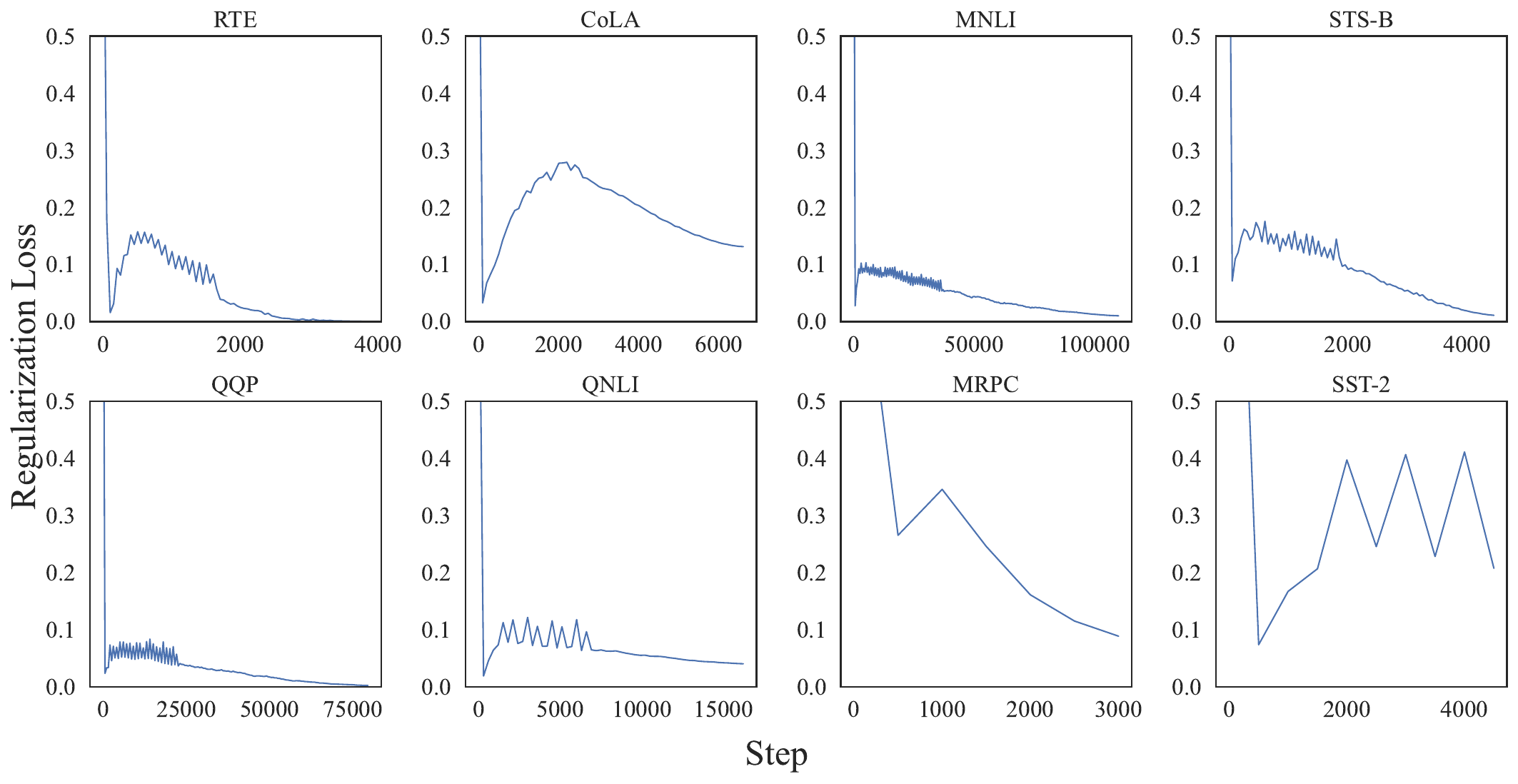}
    \caption{Orthogonal Regularization loss on all eight tasks.}
    \label{fig:regu_loss_8}
\end{figure*}

\end{appendices}